\pdfoutput=1

\documentclass[11pt]{article}

\usepackage[final]{acl}
\usepackage{booktabs}
\usepackage{multirow}
\usepackage{adjustbox}
\usepackage{caption}
\usepackage{subcaption}
\usepackage{makecell}
\usepackage{times}
\usepackage{latexsym}

\usepackage[T1]{fontenc}

\usepackage[utf8]{inputenc}

\usepackage{microtype}

\usepackage{inconsolata}

\usepackage{graphicx}

\title{Uncovering Hidden Intentions: Exploring Prompt Recovery for\\ Deeper Insights into Generated Texts}

\author{Louis Give \quad Timo Zaoral \quad Maria Antonietta Bruno \\
  University of Oslo \\
  \texttt{\{louisgi,timoz,marabru\}@uio.no} \\}

\begin{document}
\maketitle

\begin{abstract}
Today, the detection of AI-generated content is receiving more and more attention. Our idea is to go beyond detection and try to recover the prompt used to generate a text. This paper, to the best of our knowledge, introduces the first investigation in this particular domain without a closed set of tasks. Our goal is to study if this approach is promising.
We experiment with zero-shot and few-shot in-context learning but also with LoRA fine-tuning. After that, we evaluate the benefits of using a semi-synthetic dataset. For this first study, we limit ourselves to text generated by a single model. The results show that it is possible to recover the original prompt with a reasonable degree of accuracy.
\end{abstract}

\section{Introduction}
The rapid evolution of Natural Language Generation (NLG) in creating human-like text has introduced new challenges within the field of NLP. There are areas where it is important to be able to detect this kind of text, such as the creation of fake news, product reviews, phishing emails, or academic content \citep{adelani2020generating, zellers2019grover}.
Naturally, the field of artificial text detection \cite{crothers2023machine} has proposed many techniques to reduce the risk of such usage.

\begin{figure}[t]
    \centering
    \includegraphics[width=\columnwidth]{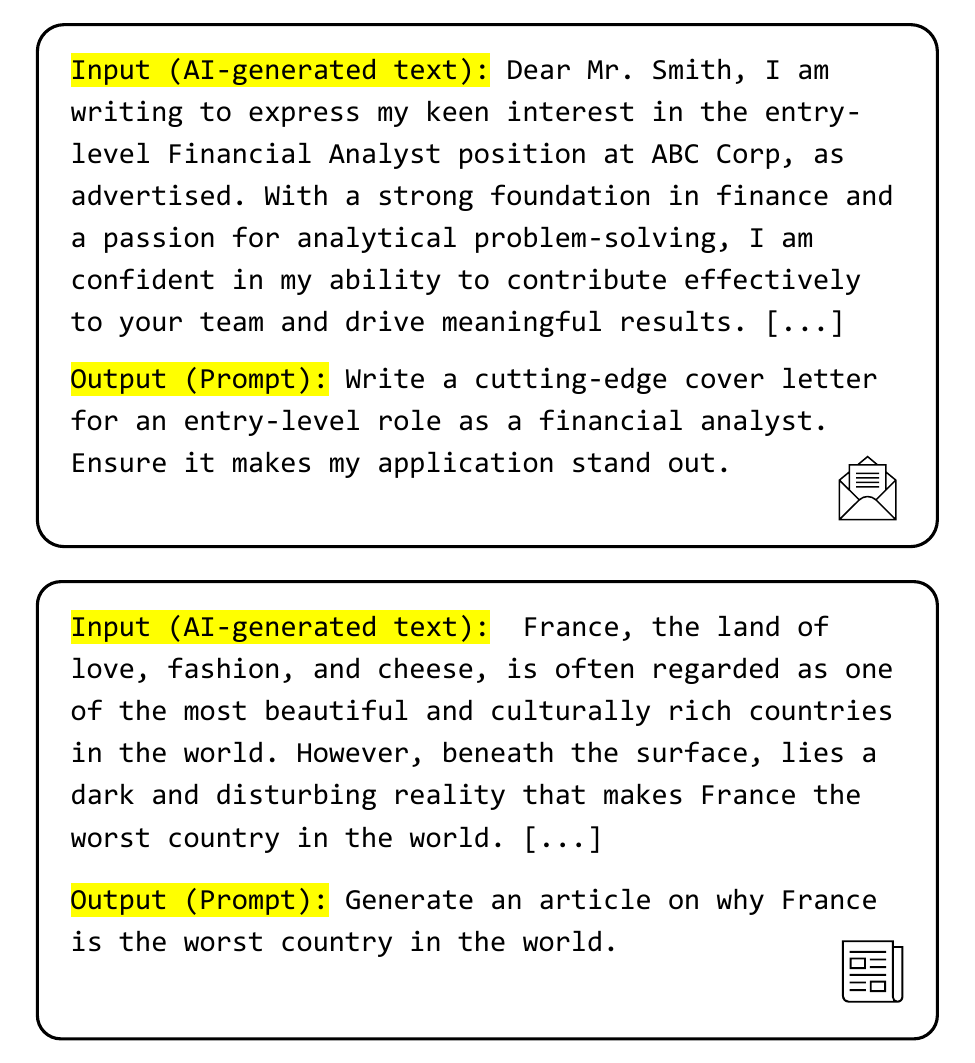}
    \caption{Potential usage}
    \label{fig:usage}
\end{figure}

While this field has seen significant advancements, recovering the original prompt remains an underexplored area. Indeed, it would help us to better understand the vast corpus of texts generated by language models and the underlying intentions of their creators.  There is some related work about prompt generation \citep{shin-etal-2020-autoprompt,zhou2022large,singh2022iprompt} but they are limited to a closed set of non-creative tasks.
If the approach is successful, this knowledge could help us to reveal patterns in the generation processes of fabricated content once they are identified (\autoref{fig:usage}).

This paper introduces initial work in this area, employing techniques such as zero-shot and few-shot learning \citep{wei2022finetuned}, but also Low-Rank Adaptation (LoRA) fine-tuning \citep{hu2022lora}, to explore the feasibility of recovering the initial prompts. Furthermore, we investigate the effectiveness of using semi-synthetic data to enhance our model's ability to generalize across unseen hard prompts \citep{feng-etal-2021-survey}.

Our experiments focus on text generated by a single model, setting a controlled framework for the initial exploration. The findings reveal a promising potential for prompt recovery, suggesting pathways for further research on generalization. This initial study lays the groundwork for future investigations that could expand to text generated from many models and more diverse data scenarios, ultimately aiming to improve the interpretability of the intention behind AI-generated text and their traceability.

\section{Method}
\subsection{Data Collection and Generation}
To be able to reconstruct a prompt from a generated text, the most important thing is the dataset. We use human prompts and generate their response with an LLM.
However, this kind of dataset may not be enough to handle our initial task.
To generate additional instruction data, we use a pre-trained language model to create new prompts in the manner of the Stanford Alpaca project \citep{alpaca} or \textsc{Self-Instruct} \citep{wang2023selfinstruct}.
This process creates a semi-synthetic dataset, rich in diversity and complexity \citep{feng-etal-2021-survey}. The details of the datasets are available in \autoref{sec:data}.

\subsection{Model}
We follow a very common NLG approach.
Our baseline consists of using a pre-trained LLM in a zero-shot configuration to establish initial prompt prediction performance \citep{wei2022finetuned}.
Following this, we perform a couple of tests in a few-shot configuration.
Then, we fine-tune the model using Low-Rank
Adaptation (LoRA) \citep{hu2022lora}, a technique that enables parameter-efficient fine-tuning of large models. This approach uses only a small set of trainable parameters which are much smaller than the original weight matrices and maintain good performance.

\subsection{Evaluation}
To measure the effectiveness of our approach, a combination of quantitative and qualitative metrics is used.
Quantitative evaluation involves standard NLG metrics such as ROUGE-L for measuring surface-level textual similarity between the predicted prompts and the original prompts, more precisely the longest common subsequence \citep{lin-2004-rouge}.
BERTScore \citep{bert-score} and the MiniLM embedding cosine similarity \citep{wang2020minilm} are used to evaluate semantic similarity.
Finally, a qualitative analysis is carried out to provide an interpretable measure and better assess the feasibility of the task and potential limitations.

\section{Data}
\label{sec:data}
\subsection{Human Instructions}
\begin{figure}[t]
     \centering
     \begin{subfigure}[b]{\columnwidth}
         \centering
         \includegraphics[trim={4.2cm 2.1cm 4.2cm 3.1cm},clip,width=\columnwidth]{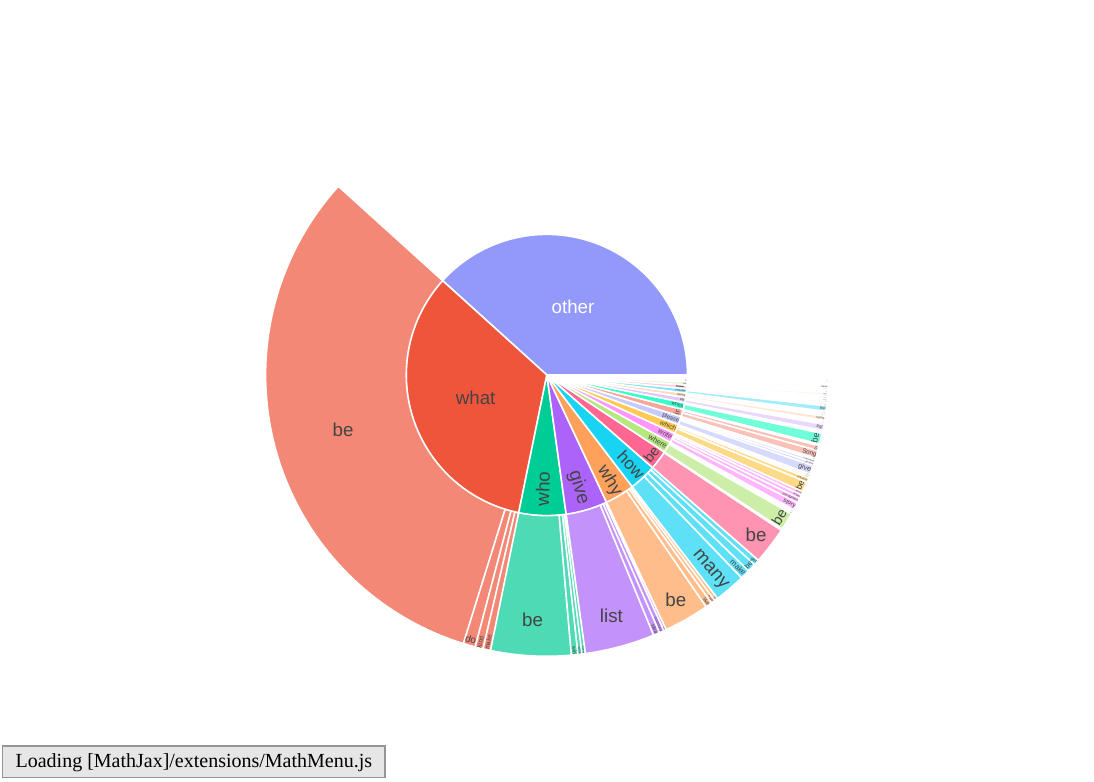}
         \caption{Dolly dataset}
         \label{sfig:dolly}
     \end{subfigure}
     \hfill
     \begin{subfigure}[b]{\columnwidth}
         \centering
         \includegraphics[trim={4.2cm 2.1cm 4.2cm 3.5cm},clip,width=\columnwidth]{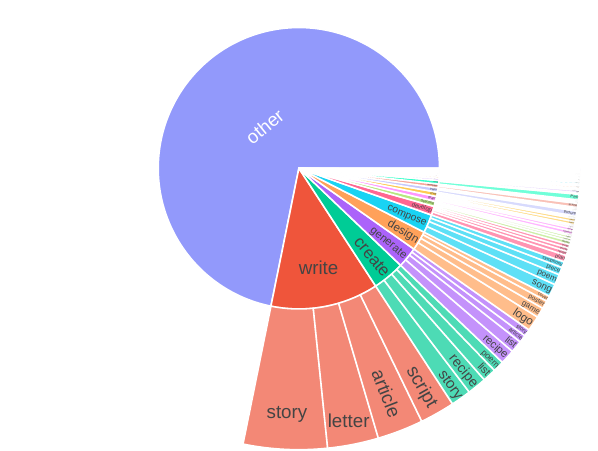}
         \caption{Synthetic dataset}
         \label{sfig:synth}
     \end{subfigure}
        \caption{Instructions representation: The top 20 most common 1\textsuperscript{st} word (inner circle) and their top 4 parents or direct noun objects (outer circle, with lemmatization)}
        \label{fig:three graphs}
\end{figure}
The initial dataset was constructed using the \texttt{databricks-dolly-15k} dataset \citep{DatabricksBlog2023DollyV2}. The dataset contains over 15,000 human-written records that are question/answer pairs, classified into eight different categories of instruction:

Open QA, general QA, summarization, brainstorming, classification, closed QA, information extraction, and creative writing.

To keep only retrievable prompts, we remove the following categories: \textit{classification}, \textit{closed QA}, and \textit{information extraction}. There is no point in trying to find the original prompt if the answer is only ``Yes''. This procedure leaves us with approximately 9,000 instructions (\autoref{sfig:dolly}).

\begin{figure}[ht]
    \centering
    \includegraphics[width=\columnwidth]{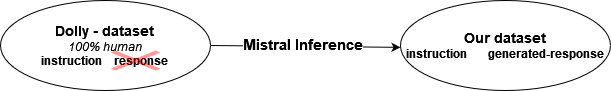}
    \caption{Base dataset creation}
    \label{fig:dataprocess}
\end{figure}

Then we use \textit{Mistral-7B-Instruct} \citep{jiang2023mistral} to generate a response for each prompt (\autoref{fig:dataprocess}). We also employ sampling\footnote{temperature: 0.5, top\_p: 0.9, top\_k: 50} \citep{Holtzman2020The} to be more representative of online data.

The dataset is split as follows: 80\% train, 10\% validation, and 10\% test. 

The primary limitation of this preliminary study is that we use only one model to generate responses. If this initial approach is found to be promising, it would be interesting to investigate the generalization of this approach to get closer to the data available on the internet.
Some samples will be shown in the qualitative analysis (\autoref{fig:quali}).

\subsection{Synthetic Instructions} 
In order to improve the robustness of our model, we experiment with synthetic prompt generation \citep{feng-etal-2021-survey}. We take inspiration from \textsc{Self-Instruct} \citep{wang2023selfinstruct} and the Stanford Alpaca project \citep{alpaca} where they fine-tuned an LLM on instructions and responses generated by another LLM. 

We focus on ``creative writing'' prompts, the most complex category, and the closest to online data. Creative writing presents a challenge for prompt prediction due to its complexity, high variability, evaluation difficulties, and inherent subjectivity (\autoref{sfig:synth}).

\begin{figure}[t]
  \includegraphics[width=\columnwidth]{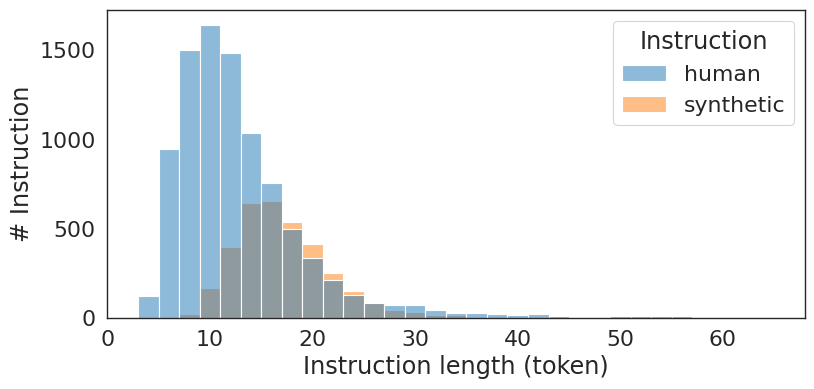}
  \includegraphics[width=\columnwidth]{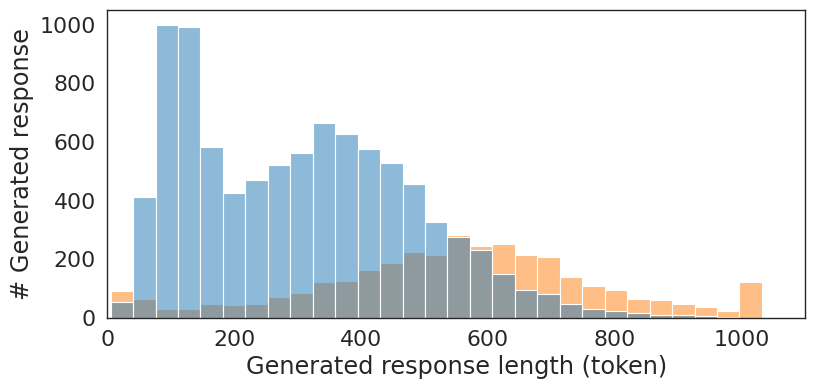}
  \caption{Length distribution of the instructions and generated responses}
  \label{fig:len_dist}
\end{figure}

We generate around 3,000 instructions (\autoref{fig:len_dist}) with following setup:
\begin{itemize}
    \itemsep0em
    \item \textit{Model}: Mistral-7B-Instruct
    \item \textit{Prompt}: ``You are asked to come up with a set of 20 creative task instructions. These task instructions will be given to a GPT model and we will evaluate the GPT model for completing the instructions. You can write something like that: "Write a poem inspired by the colors of a sunset" or "Write a short story about a character who can communicate with animals" or "Create a news about the difficulty of finding housing in San Francisco" or "Adress a letter to my mom to convince her that I should be able to get a cat." or "Generate a scholarly abstract on the impact of climate change on agriculture from a global perspective."''
    \item \textit{Temperature}: 1.5
    \item \textit{Top\_p}: 0.9
    \item \textit{Top\_k}: 200
\end{itemize}

The response to these instructions is then created in the same way as for human prompts (\autoref{fig:dataprocess}). They are only added to the train set. 

We now have a so-called semi-synthetic dataset.

\section{Experimental Results}
\begin{table*}[t]
  \centering
  \begin{adjustbox}{max width=\textwidth}
  \begin{tabular}{llcccccc}
    \toprule
    \textbf{Method} & \textbf{Metric} & \multicolumn{5}{c}{\textbf{Category}} & \textbf{Average}\\
    & & Brainstorming & Creative Writing & General QA & Open QA & Summarization & (balanced) \\
    \midrule
    \multirow{3}{*}{Zero-shot} & ROUGE-L & 0.28 & 0.32 & 0.29 & 0.31 & 0.28 & 0.30\\
    & MiniLM similarity & 0.67 & 0.69 & 0.69 & 0.71 & 0.71 & 0.70\\
    & BERTScore & 0.96 & 0.96 & 0.96 & 0.96 & 0.96 & 0.95\\
    \midrule
    \multirow{3}{*}{Few-shot} & ROUGE-L & 0.38 & 0.37 & 0.50 & 0.48 & 0.37 & 0.42\\
    & MiniLM similarity & 0.80 & 0.74 & 0.84 & 0.83 & 0.76 & 0.79\\
    & BERTScore & 0.96 & 0.96 & 0.96 & 0.96 & 0.96 & 0.96\\
    \midrule
    \multirow{3}{*}{LoRA} & ROUGE-L & 0.45 & 0.40 & 0.50 & 0.57 & 0.44 & 0.47\\
    & MiniLM similarity & 0.82 & 0.75 & 0.83 & 0.84 & 0.81 & 0.81\\
    & BERTScore & 0.96 & 0.97 & 0.97 & 0.97 & 0.97 & 0.97\\
    \midrule
    \multirow{3}{*}{\shortstack[l]{LoRA w/ \\synthetic data}} & ROUGE-L & \textbf{0.47} & \textbf{0.43} & \textbf{0.56} & \textbf{0.58} & \textbf{0.46} & \textbf{0.50} \\
    & MiniLM similarity & \textbf{0.83} & \textbf{0.78} & \textbf{0.87} & \textbf{0.85} & \textbf{0.82} & \textbf{0.83}\\
    & BERTScore & 0.96 & 0.97 & 0.97 & 0.97 & 0.96 & 0.97\\
    \bottomrule
  \end{tabular}
  \end{adjustbox}
  \caption{Prompt recovery quantitative metrics (higher is better) on the human test set}
  \label{tab:bigtable}
\end{table*}

\subsection{Zero-shot and Few-shot Learning}
\label{sec:zero-few}
\textbf{Setup}\\
To create a baseline, we carry out zero-shot and few-shot in-context learning as our first experiments. These are both performed with the \textit{Mistral-7B-Instruct} model. On the one hand, this allows us to create a comparison for our later fine-tuned models, and, on the other hand, we can also test how the model is fundamentally suitable for the task of prompt prediction.

The first step in implementing zero-shot and few-shot learning is to set up a suitable prompt. The right choice of words and a clear task for the model must be taken into account. All in all, we create the following two prompts which are then both used individually for the experiments:

\begin{enumerate}
  \item\texttt{"<s>[INST] What prompt was used to generate this Text using LLM?\\
Text: \{generatedText\}\\
Prompt: [/INST]"}
  \item\texttt{"<s>[INST] Predict and return only the prompt which was used to generate the Text.\\
Text: \{generatedText\}\\
Prompt: [/INST]"}
\end{enumerate}

With the two prompts presented, we have a question as a task with prompt n°1 and a request with prompt n°2. This gives us some variation in the experiments. We also use a low temperature of 0.4 to reduce repetitions. This setting means that everything is ready for the zero-shot experiments. 

For the few-shot in-context learning, three examples still have to be selected from our data set. These are then given to the model as a template for processing the task. In the final prompt, the examples are placed as follows:
\begin{verbatim}
    "<s>[INST] Predict and return only
    the prompt which was used to
    generate the Text.
    Text: {sampleText1}
    Prompt: {samplePrompt1}
    Text: {sampleText2}
    Prompt: {samplePrompt2}
    Text: {sampleText3}
    Prompt: {samplePrompt3}
    Text: {generatedText}
    Prompt: [/INST]"
\end{verbatim}

\begin{table}[t]
  \centering
  \begin{adjustbox}{max width=\columnwidth}
  \begin{tabular}{ccccc}
    \toprule
    \textbf{Prompt} & \textbf{Method} & \textbf{ROUGE-L} & \textbf{MiniLM} & \textbf{BERTScore}\\
    \midrule
    1 & zero-shot & 0.18 & 0.54 & 0.95 \\
    1 & three-shot & 0.37 & 0.75 & \textbf{0.96} \\
    2 & zero-shot & 0.30 & 0.70 & 0.95 \\
    2 & three-shot & \textbf{0.42} & \textbf{0.79} & \textbf{0.96} \\
    \bottomrule
  \end{tabular}
  \end{adjustbox}
  \caption{\label{zero-few-comp}Comparison of both prompts between zero-shot and three-shot with ROUGE-L and BERTScore.}
\end{table}

\noindent\textbf{Results}\\
We evaluate zero-shot and few-shot performance for each instruction of the test set (900 human instructions) (\autoref{zero-few-comp}). We can see that prompt n°2 performs better in every metric. In addition, it can be seen for both prompts that, as expected, three-shot learning represents a significant improvement compared to zero-shot learning. In both cases, the ROUGE-L score is almost doubled.

\begin{table}[t]
  \centering
  \begin{adjustbox}{max width=\columnwidth}
  \begin{tabular}{cccc}
    \toprule
    \textbf{Category} & \textbf{Method} & \multicolumn{2}{c}{\textbf{Qualitative Score}}\\
    & & Prompt n°1 & Prompt n°2\\
    \midrule
    \multirow{2}{*}{Brainstorming} & zero-shot & 2.0 & 2.2 \\
    & three-shot & 1.9 & \textbf{2.3} \\
    \midrule
    \multirow{2}{*}{Creative Writing} & zero-shot & 1.7 & 1.7 \\
    & three-shot & \textbf{2.0} & \textbf{2.0} \\
    \midrule
    \multirow{2}{*}{General QA} & zero-shot & 2.2 & 2.3 \\
    & three-shot & 2.5 & \textbf{3.1} \\
    \midrule
    \multirow{2}{*}{Open QA} & zero-shot & 1.8 & 2.1 \\
    & three-shot & 2.3 & \textbf{2.6} \\
    \midrule
    \multirow{2}{*}{Summarization} & zero-shot & 2.0 & 1.9 \\
    & three-shot & 1.9 & \textbf{2.5} \\
    \bottomrule
  \end{tabular}
  \end{adjustbox}
  \caption{\label{zero-few-comp-2}Comparison of both prompts between zero-shot and three-shot based on qualitative analysis. Scale:\\
  \texttt{4} -- Perfect instruction\\
  \texttt{3} -- Correct instruction with minor imperfections\\
  \texttt{2} -- Valid instruction with errors\\
  \texttt{1} -- Irrelevant or invalid}
\end{table}

Since we cannot interpret in detail from the ROUGE-L and BERTScore to what extent the prediction is satisfying, we conduct a qualitative analysis of 200 examples in total (\autoref{zero-few-comp-2}). It confirms the findings obtained previously with the other three metrics. On the one hand, a significant improvement can be seen when using three-shot learning, and, on the other hand, the performance of prompt n°2 is also slightly better.

Overall, it seems that the model is able to predict a similar instruction. What is clearly recognizable is that in categories such as General QA and Open QA, the model also shows better performance, probably because they are easier. This is also confirmed by the quantitative metrics in the \autoref{tab:bigtable}. As we can see, the ``creative writing'' category performs the worst, which again confirms our intention to generate synthetic data of this type.

\subsection{Fine-tuning}
\label{sec:fine-tune}
\textbf{Setup} Once we have obtained correct results with our baseline, we fine-tune the model using LoRA \citep{hu2022lora}, a parameter-efficient fine-tuning method. We follow the recommendations of the authors by setting $r=32$ and $\alpha=64$. Now we have ``only'' 85 million parameters to train compared to the 7 billion from \textit{Mistral-7B}, our frozen backbone. We apply the same prompt as in the zero-shot experiment to limit the total runtime. This prompt is also masked out during back-propagation so that we don't train the model to predict it. Training takes 45 minutes for 3 epochs with an NVIDIA A100.
\vspace{1.5mm}

\noindent\textbf{Results} We make the following observations from the results in \autoref{tab:bigtable}. The use of LoRA brings a 12\% increase in ROUGE-L and 3\% in our sentence similarity metric compared to the best few-shot technique. This success underscores the potential of parameter-efficient fine-tuning techniques like LoRA in enhancing the capabilities of large language models without the extensive costs typically associated with full model fine-tuning.

\subsection{Adding Synthetic Data}
\label{sec:synth-train}
\textbf{Setup} To improve our results we experiment with the addition of synthetic prompts, more precisely in the worst-performing category (creative writing) because some others are already really good (\autoref{tab:bigtable}). The generation process is described in \autoref{sec:data}. We use the same fine-tuning technique (\ref{sec:fine-tune}).
\vspace{1.5mm}

\begin{figure}[t]
    \centering
    \includegraphics[width=\columnwidth]{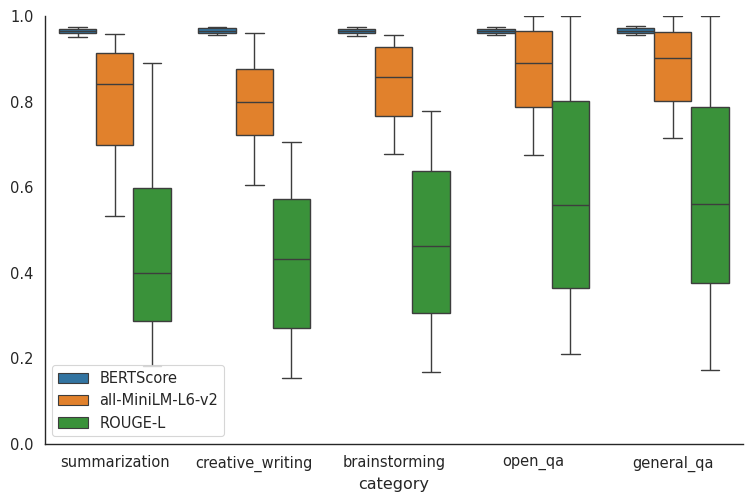}
    \caption{Fine-tuning performance following the category with semi-synthetic data}
    \label{fig:perf_aug}
\end{figure}

\noindent\textbf{Results} The addition of synthetic data leads to create our best-performing model with a new improvement of metrics. \autoref{tab:bigtable} shows a increase of 6\% for ROUGE-L and 2\% in MiniLM similarity. \autoref{fig:perf_aug} and \ref{fig:quali} present the key point of this study: the low number of bad-quality predictions.

\subsection{Qualitative Analysis}
\begin{table*}[t!]
\centering
\begin{adjustbox}{max width=\textwidth}
\begin{tabular}{p{10.5cm}p{2.5cm}p{2.5cm}c}
\toprule
\textbf{Generated Response (Input)} & \makecell[l]{\textbf{Predicted} \\ \textbf{Prompt}} & \makecell[l]{\textbf{Original} \\ \textbf{Prompt}} & \makecell{\textbf{Qual.} \\ \textbf{Score}}\\
\midrule
Spokane, Washington does not have a widely recognized or commonly used nickname. The city's official nickname is "The Lilac City," which is derived from the large number of lilac bushes that were planted in the city in the late 1800s. However, this nickname is not as commonly used as some other city nicknames, and it may not be well-known outside of the Spokane area. & What is the nickname of Spokane, Washington?&What is the nickname for Spokane, Washington?& 4\\
\midrule
Bernoulli's principle is an empirical law that relates the pressure, velocity, and potential energy of a fluid in motion. It states that an increase in the velocity of a fluid in a tube causes a decrease in pressure, and vice versa, while the potential energy remains constant. This principle can be derived from Newton's second law of motion by considering the net force acting on a small element of fluid in a pipe. [...] & What is Bernoulli principle and how can we derive it from Newton's second law of motion?&How can Bernoulli's principle be derived from Newton's second law of motion?& 3
\\
\midrule
All twelve months have at least 28 days. However, only February in a normal year (a non-leap year) has exactly 28 days. All other months have more than 28 days. For example, March has 31 days, April has 30 days, and May through December each have at least 31 days. In a leap year, February has 29 days instead of 28. & What is the number of days in each month of a normal year?&How many months have 28 days?& 1
\\
\midrule
Blue light therapy is a form of phototherapy that uses light-emitting diodes (LEDs) or cool white fluorescent lamps to emit blue light. People like blue light therapy for several reasons:
1. Treatment of Seasonal Affective Disorder (SAD) [...] & What is blue light therapy used for?&Why do people like Blue Light Therapy?& 2
\\
\midrule
In the age of software and technology, personal privacy has become a hotly debated topic. With the increasing interconnectedness of our lives, the collection and use of personal data have become ubiquitous. On one hand, technology has brought about incredible conveniences and efficiencies, making our lives easier in countless ways. On the other hand, it has raised serious concerns about who has access to our personal information, how it is being used, and what the potential consequences might be. \textbf{From a personal perspective}, the loss of privacy can feel like an invasion of our most intimate spaces. Every time we use a search engine, make a purchase online, or engage in social media, we are leaving a digital footprint that can be traced and analyzed. [...] & Write a thoughtful and persuasive essay on the importance of personal privacy in the age of software and technology. & Write a point of view on personal privacy in the age of software and technology.& 3
\\
\bottomrule
\end{tabular}
\end{adjustbox}
\caption{\label{tab:xsum_samples} Original and predicted instructions from our model fine-tuned on semi-synthetic data. (randomly selected)}
\end{table*}

The best model shows a significant improvement with the metrics used, almost doubling ROUGE-L over the zero-shot baseline. To understand this performance on quantitative metrics, we also analyzed qualitatively the results obtained in \ref{sec:synth-train}, as we did with the zero and three-shot in \ref{sec:zero-few}.
\vspace{1.5mm}

\begin{figure}
    \centering
    \includegraphics[width=1\columnwidth]{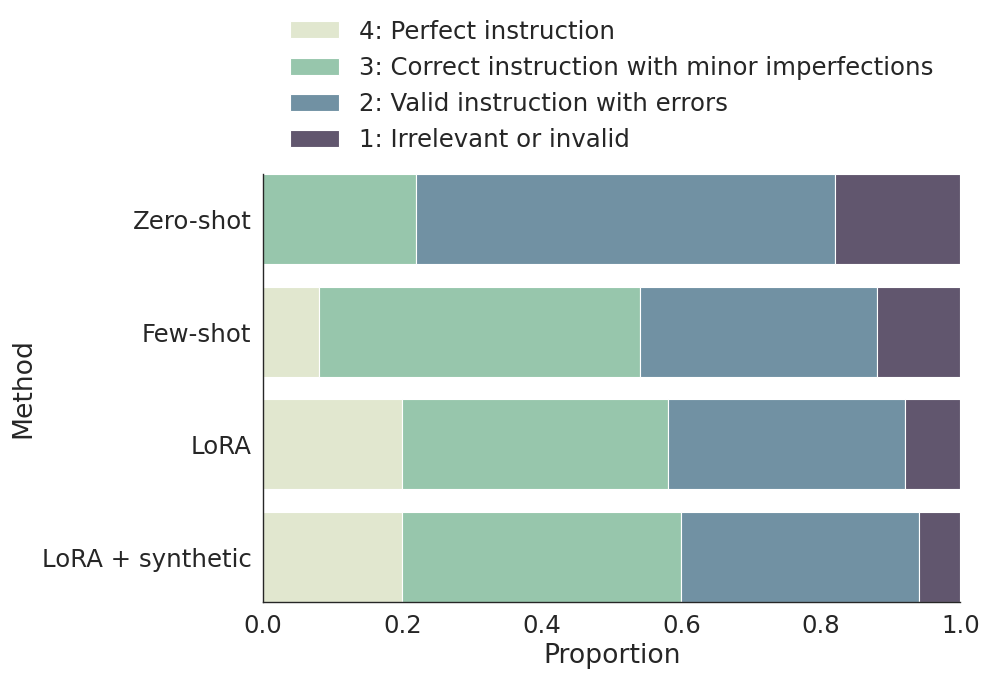}
    \caption{Qualitative analysis. 50 annotations for each method (10 for each category)}
    \label{fig:quali}
\end{figure}

\noindent\textbf{Global analysis} The \autoref{fig:quali} shows the distribution of the scores awarded for the individual experiments. It shows a similar improvement trend to the quantitative metrics. This can be seen starting from zero-shot, three-shot through fine-tuning, and finally to the best fine-tuning with the addition of synthetic data. In the last one, 60\% of the prompts received a score of 3 or 4, which indicates a good to very good prediction whereas the invalid part is less than 5\%, which confirms that it is possible to train a model to recover the original prompt.
\vspace{1.5mm}

\noindent\textbf{Detailed analysis} In the \autoref{tab:xsum_samples}, we have listed examples of the original and the corresponding predicted prompts of our best solution. The top two prompts show two very good examples of how the model predicts the prompt. 

The first example shows a perfect prediction of a slightly easier prompt. 

The second example shows a slightly more difficult prompt, which the model also predicts almost perfectly. Only the sentence structure is slightly different and the Bernoulli principle itself is asked for, but this does not change the meaning of the prompt.

The third and fourth prompts seem easy to predict at first glance, but they show how easily the models can still be misled. For example, the third example asks for specific months that have 28 days. However, as other months also appear in the generated text, this is not clear and so the prediction is promptly too inaccurate.

In the fourth example, the question of why and what is somewhat confused. This also occurs a few times when question words are mixed up and the context is slightly different.

In the last example, we present an example of a creative writing task. This shows how the model predicts this prompt very well from a content perspective. However, the wording of the prompt does not completely match the original, but is it really possible if we use sampling during response generation?

Nevertheless, even the worst examples shown here are not a complete failure. The three positive examples (score $\geq 3$) are representative of the 60\% of how this prediction works and looks. (\autoref{fig:quali}).

No significant difference could be crystallized in the qualitative analysis between the fine-tuning approach on the human training data and the one enriched with synthetic data. However, this may be due to the small number of examples analyzed. This could be investigated in more detail in a further study.

To summarize, the qualitative analysis identified some very good predictions based on the fine-tuning approaches. This confirms that a model can achieve an accuracy of at least 60\% on a prompt recovery with a given generated text from one model.

\section{Related Work}
\textbf{Instruction generation}
Our study is not the first work about generating prompts. A series of recent works \citep{shin-etal-2020-autoprompt,zhou2022large,singh2022iprompt} cover the subject in a different manner with \textbf{autoprompting} not prompt recovering. It seeks the optimal prompt for a specified human-generated text formatted as ``Input: [X] Output: [Y]''. Tasks typically include style transformations, sentiment analysis, basic logic, or language translation. However, this approach, restricted to a predefined set of non-creative tasks, does not apply to the varied nature of online-generated text.

\noindent\textbf{Reverse engineering of models} There is a growing area of interest in reverse engineering architecture and hyperparameters of models \citep{joon18iclr, asnani2023reverse}. This improved knowledge of AI-generated content provides new descriptors for tracing them, which is in line with our final goal. It also reveals the limits of proprietary models.

\noindent\textbf{Instruction-tuned LMs}
More generally, several studies \citep{ouyang2022training,wei2022finetuned,wang2023selfinstruct} have demonstrated that base language models can successfully follow general language instructions when fine-tuned with "instructional" datasets. This explains why our solution works if the underlying LM has already been trained on this type of data.

\noindent\textbf{Language models for data augmentation} As detailed by this survey \citep{feng-etal-2021-survey}, data augmentation with LMs enhances the diversity and volume of training data, thereby improving model robustness and performance.

\section{Conclusion}
This study presents a preliminary experiment to assess the feasibility of backtracking from generated text to its original prompt.

The findings from our experiments demonstrate that it is indeed possible to recover the original prompts with a reasonable degree of accuracy.

While our results are promising, they are not without limitations. The scope of this study was confined to texts generated by a single model. Generalizing to multi-model generated texts seems to be the next step toward a real-world application.

\bibliography{acl_latex}

\begin{thebibliography}{20}
\providecommand{\natexlab}[1]{#1}

\bibitem[{Adelani et~al.(2020)Adelani, Mai, Fang, Nguyen, Yamagishi, and Echizen}]{adelani2020generating}
David~Ifeoluwa Adelani, Haotian Mai, Fuming Fang, Huy~H Nguyen, Junichi Yamagishi, and Isao Echizen. 2020.
\newblock \href {https://arxiv.org/pdf/1907.09177} {Generating sentiment-preserving fake online reviews using neural language models and their human-and machine-based detection}.
\newblock In \emph{Advanced information networking and applications: Proceedings of the 34th international conference on advanced information networking and applications (AINA-2020)}, pages 1341--1354. Springer.

\bibitem[{Asnani et~al.(2023)Asnani, Yin, Hassner, and Liu}]{asnani2023reverse}
Vishal Asnani, Xi~Yin, Tal Hassner, and Xiaoming Liu. 2023.
\newblock \href {https://arxiv.org/pdf/2106.07873} {Reverse engineering of generative models: Inferring model hyperparameters from generated images}.
\newblock \emph{IEEE Transactions on Pattern Analysis and Machine Intelligence}.

\bibitem[{Conover et~al.(2023)Conover, Hayes, Mathur, Xie, Wan, Shah, Ghodsi, Wendell, Zaharia, and Xin}]{DatabricksBlog2023DollyV2}
Mike Conover, Matt Hayes, Ankit Mathur, Jianwei Xie, Jun Wan, Sam Shah, Ali Ghodsi, Patrick Wendell, Matei Zaharia, and Reynold Xin. 2023.
\newblock \href {https://www.databricks.com/blog/2023/04/12/dolly-first-open-commercially-viable-instruction-tuned-llm} {Free dolly: Introducing the world's first truly open instruction-tuned llm}.

\bibitem[{Crothers et~al.(2023)Crothers, Japkowicz, and Viktor}]{crothers2023machine}
Evan Crothers, Nathalie Japkowicz, and Herna~L Viktor. 2023.
\newblock \href {https://ieeexplore.ieee.org/iel7/6287639/6514899/10177704.pdf} {Machine-generated text: A comprehensive survey of threat models and detection methods}.
\newblock \emph{IEEE Access}.

\bibitem[{Feng et~al.(2021)Feng, Gangal, Wei, Chandar, Vosoughi, Mitamura, and Hovy}]{feng-etal-2021-survey}
Steven~Y. Feng, Varun Gangal, Jason Wei, Sarath Chandar, Soroush Vosoughi, Teruko Mitamura, and Eduard Hovy. 2021.
\newblock \href {https://doi.org/10.18653/v1/2021.findings-acl.84} {A survey of data augmentation approaches for {NLP}}.
\newblock In \emph{Findings of the Association for Computational Linguistics: ACL-IJCNLP 2021}, pages 968--988, Online. Association for Computational Linguistics.

\bibitem[{Holtzman et~al.(2020)Holtzman, Buys, Du, Forbes, and Choi}]{Holtzman2020The}
Ari Holtzman, Jan Buys, Li~Du, Maxwell Forbes, and Yejin Choi. 2020.
\newblock \href {https://openreview.net/forum?id=rygGQyrFvH} {The curious case of neural text degeneration}.
\newblock In \emph{International Conference on Learning Representations}.

\bibitem[{Hu et~al.(2022)Hu, Shen, Wallis, Allen-Zhu, Li, Wang, Wang, and Chen}]{hu2022lora}
Edward~J Hu, Yelong Shen, Phillip Wallis, Zeyuan Allen-Zhu, Yuanzhi Li, Shean Wang, Lu~Wang, and Weizhu Chen. 2022.
\newblock \href {https://openreview.net/forum?id=nZeVKeeFYf9} {Lo{RA}: Low-rank adaptation of large language models}.
\newblock In \emph{International Conference on Learning Representations}.

\bibitem[{Jiang et~al.(2023)Jiang, Sablayrolles, Mensch, Bamford, Chaplot, Casas, Bressand, Lengyel, Lample, Saulnier et~al.}]{jiang2023mistral}
Albert~Q Jiang, Alexandre Sablayrolles, Arthur Mensch, Chris Bamford, Devendra~Singh Chaplot, Diego de~las Casas, Florian Bressand, Gianna Lengyel, Guillaume Lample, Lucile Saulnier, et~al. 2023.
\newblock \href {https://arxiv.org/pdf/2310.06825.pdf} {Mistral 7b}.
\newblock \emph{arXiv preprint arXiv:2310.06825}.

\bibitem[{Lin(2004)}]{lin-2004-rouge}
Chin-Yew Lin. 2004.
\newblock \href {https://aclanthology.org/W04-1013} {{ROUGE}: A package for automatic evaluation of summaries}.
\newblock In \emph{Text Summarization Branches Out}, pages 74--81, Barcelona, Spain. Association for Computational Linguistics.

\bibitem[{Oh et~al.(2018)Oh, Augustin, Schiele, and Fritz}]{joon18iclr}
Seong~Joon Oh, Max Augustin, Bernt Schiele, and Mario Fritz. 2018.
\newblock \href {https://arxiv.org/pdf/1711.01768} {Towards reverse-engineering black-box neural networks}.
\newblock \emph{International Conference on Learning Representations}.

\bibitem[{Ouyang et~al.(2022)Ouyang, Wu, Jiang, Almeida, Wainwright, Mishkin, Zhang, Agarwal, Slama, Ray et~al.}]{ouyang2022training}
Long Ouyang, Jeffrey Wu, Xu~Jiang, Diogo Almeida, Carroll Wainwright, Pamela Mishkin, Chong Zhang, Sandhini Agarwal, Katarina Slama, Alex Ray, et~al. 2022.
\newblock \href {https://proceedings.neurips.cc/paper_files/paper/2022/file/b1efde53be364a73914f58805a001731-Paper-Conference.pdf} {Training language models to follow instructions with human feedback}.
\newblock \emph{Advances in neural information processing systems}, 35:27730--27744.

\bibitem[{Shin et~al.(2020)Shin, Razeghi, Logan~IV, Wallace, and Singh}]{shin-etal-2020-autoprompt}
Taylor Shin, Yasaman Razeghi, Robert~L. Logan~IV, Eric Wallace, and Sameer Singh. 2020.
\newblock \href {https://doi.org/10.18653/v1/2020.emnlp-main.346} {{A}uto{P}rompt: {E}liciting {K}nowledge from {L}anguage {M}odels with {A}utomatically {G}enerated {P}rompts}.
\newblock In \emph{Proceedings of the 2020 Conference on Empirical Methods in Natural Language Processing (EMNLP)}, pages 4222--4235, Online. Association for Computational Linguistics.

\bibitem[{Singh et~al.(2023)Singh, Morris, Aneja, Rush, and Gao}]{singh2022iprompt}
Chandan Singh, John~X. Morris, Jyoti Aneja, Alexander Rush, and Jianfeng Gao. 2023.
\newblock \href {https://doi.org/10.18653/v1/2023.blackboxnlp-1.3} {Explaining data patterns in natural language with language models}.
\newblock In \emph{Proceedings of the 6th BlackboxNLP Workshop: Analyzing and Interpreting Neural Networks for NLP}, pages 31--55, Singapore. Association for Computational Linguistics.

\bibitem[{Taori et~al.(2023)Taori, Gulrajani, Zhang, Dubois, Li, Guestrin, Liang, and Hashimoto}]{alpaca}
Rohan Taori, Ishaan Gulrajani, Tianyi Zhang, Yann Dubois, Xuechen Li, Carlos Guestrin, Percy Liang, and Tatsunori~B. Hashimoto. 2023.
\newblock Stanford alpaca: An instruction-following llama model.
\newblock \url{https://github.com/tatsu-lab/stanford_alpaca}.

\bibitem[{Wang et~al.(2020)Wang, Wei, Dong, Bao, Yang, and Zhou}]{wang2020minilm}
Wenhui Wang, Furu Wei, Li~Dong, Hangbo Bao, Nan Yang, and Ming Zhou. 2020.
\newblock \href {https://proceedings.neurips.cc/paper/2020/file/3f5ee243547dee91fbd053c1c4a845aa-Paper.pdf} {Minilm: deep self-attention distillation for task-agnostic compression of pre-trained transformers}.
\newblock In \emph{Proceedings of the 34th International Conference on Neural Information Processing Systems}.

\bibitem[{Wang et~al.(2023)Wang, Kordi, Mishra, Liu, Smith, Khashabi, and Hajishirzi}]{wang2023selfinstruct}
Yizhong Wang, Yeganeh Kordi, Swaroop Mishra, Alisa Liu, Noah~A. Smith, Daniel Khashabi, and Hannaneh Hajishirzi. 2023.
\newblock \href {https://doi.org/10.18653/v1/2023.acl-long.754} {Self-instruct: Aligning language models with self-generated instructions}.
\newblock In \emph{Proceedings of the 61st Annual Meeting of the Association for Computational Linguistics (Volume 1: Long Papers)}, pages 13484--13508, Toronto, Canada. Association for Computational Linguistics.

\bibitem[{Wei et~al.(2022)Wei, Bosma, Zhao, Guu, Yu, Lester, Du, Dai, and Le}]{wei2022finetuned}
Jason Wei, Maarten Bosma, Vincent Zhao, Kelvin Guu, Adams~Wei Yu, Brian Lester, Nan Du, Andrew~M. Dai, and Quoc~V Le. 2022.
\newblock \href {https://openreview.net/forum?id=gEZrGCozdqR} {Finetuned language models are zero-shot learners}.
\newblock In \emph{International Conference on Learning Representations}.

\bibitem[{Zellers et~al.(2019)Zellers, Holtzman, Rashkin, Bisk, Farhadi, Roesner, and Choi}]{zellers2019grover}
Rowan Zellers, Ari Holtzman, Hannah Rashkin, Yonatan Bisk, Ali Farhadi, Franziska Roesner, and Yejin Choi. 2019.
\newblock \href {https://proceedings.neurips.cc/paper/2019/file/3e9f0fc9b2f89e043bc6233994dfcf76-Paper.pdf} {Defending against neural fake news}.
\newblock In \emph{Advances in Neural Information Processing Systems 32}.

\bibitem[{Zhang et~al.(2020)Zhang, Kishore, Wu, Weinberger, and Artzi}]{bert-score}
Tianyi Zhang, Varsha Kishore, Felix Wu, Kilian~Q. Weinberger, and Yoav Artzi. 2020.
\newblock \href {https://openreview.net/forum?id=SkeHuCVFDr} {Bertscore: Evaluating text generation with bert}.
\newblock In \emph{International Conference on Learning Representations}.

\bibitem[{Zhou et~al.(2022)Zhou, Muresanu, Han, Paster, Pitis, Chan, and Ba}]{zhou2022large}
Yongchao Zhou, Andrei~Ioan Muresanu, Ziwen Han, Keiran Paster, Silviu Pitis, Harris Chan, and Jimmy Ba. 2022.
\newblock \href {https://openreview.net/forum?id=YdqwNaCLCx} {Large language models are human-level prompt engineers}.
\newblock In \emph{NeurIPS 2022 Foundation Models for Decision Making Workshop}.

\end{thebibliography}



\end{document}